\pdfoutput=1

\documentclass[11pt]{article}

\usepackage{EMNLP2022}

\usepackage{times}
\usepackage{latexsym}

\usepackage[T1]{fontenc}

\usepackage[utf8]{inputenc}

\usepackage{microtype}

\usepackage{amsmath,amsfonts,amssymb}
\usepackage{graphicx}
\usepackage{booktabs}
\usepackage{multirow}
\usepackage[referable]{threeparttablex}
\usepackage{xspace}
\usepackage{tablefootnote}

\title{Salient Phrase Aware Dense Retrieval:\\Can a Dense Retriever Imitate a Sparse One?}

\author{
Xilun Chen, Kushal Lakhotia\textsuperscript{\textdagger}, Barlas Oğuz, Anchit Gupta,\\
\textbf{Patrick Lewis, Stan Peshterliev, Yashar Mehdad, Sonal Gupta} and \textbf{Wen-tau Yih}\\
Meta AI\\
\scalebox{0.88}{\texttt{\{xilun,barlaso,anchit,plewis,stanvp,mehdad,sonalgupta,scottyih\}@meta.com}}\\
\scalebox{0.88}{\textsuperscript{\textdagger}\texttt{lakhotia.kushal@gmail.com}}
}

\newcommand{\spar}{\texttt{SPAR}\xspace}
\newcommand{\lexmodel}{Lexical Model}
\newcommand{\lexmodelsymbol}{$\bf \Lambda$\xspace}

\newcommand{\secref}[1]{\S\ref{#1}}
\newcommand{\expnumber}[2]{{#1}\mathrm{e}{#2}}
\DeclareMathOperator{\score}{sim}

\begin{document}
\maketitle
\begin{abstract}

Despite their recent popularity and well-known advantages, dense retrievers still lag behind sparse methods such as BM25 in their ability to reliably match salient phrases and rare entities in the query and to generalize to out-of-domain data. 
It has been argued that this is an inherent limitation of dense models. 
We rebut this claim by introducing the Salient Phrase Aware Retriever (\spar{})\footnote{The code and models of \spar{} are available at: 
\url{https://github.com/facebookresearch/dpr-scale/tree/main/spar}.}, 
a dense retriever with the lexical matching capacity of a sparse model. 
We show that \emph{a dense \lexmodel{} \lexmodelsymbol{} can be trained to imitate a sparse one}, and \spar{} is built by augmenting a standard dense retriever with \lexmodelsymbol{}.
Empirically, \spar shows superior performance on a range of tasks including five question answering datasets, MS MARCO passage retrieval, as well as the EntityQuestions and BEIR benchmarks for \emph{out-of-domain} evaluation, exceeding the performance of state-of-the-art dense and sparse retrievers.

\end{abstract}


\section{Introduction}\label{sec:intro}
Text retrieval is a crucial component for a wide range of knowledge-intensive NLP systems, such as open-domain question answering (ODQA) models and search engines.
Recently, dense retrievers~\cite{karpukhin-etal-2020-dense,xiong2021approximate} have gained popularity and demonstrated strong performance on a number of retrieval tasks.
Dense retrievers employ deep neural networks to learn continuous representations for the queries and documents, and perform retrieval in this dense embedding space using nearest neighbor search~\cite{faiss}.
Compared to traditional sparse retrievers that rely on discrete bag-of-words representations, dense retrievers can derive more semantically expressive embeddings, thanks to its end-to-end learnability and powerful pre-trained encoders.
This helps dense retrievers to overcome several inherent limitations of sparse systems such as \emph{vocabulary mismatch} (where different words are used for the same meaning) and \emph{semantic mismatch} (where the same word has multiple meanings).

On the other hand, while existing dense retrievers excel at capturing semantics, they sometimes fail to match the \emph{salient phrases} in the query.
For example, \citet{karpukhin-etal-2020-dense} show that DPR, unlike a sparse BM25 retriever~\cite{bm25}, is unable to catch the salient phrase ``\emph{Thoros of Myr}'' in the query ``\emph{Who plays Thoros of Myr in Game of Thrones?}''.
In addition, dense retrievers struggle to \emph{generalize to out-of-domain test data} compared to training-free sparse retrievers such as BM25.
For instance, \citet{sciavolino2021simple} find that DPR performs poorly compared to BM25 on simple entity-centric questions, and \citet{thakur2021beir} introduce a new BEIR benchmark to evaluate the zero-shot generalization of retrieval models showing that BM25 outperforms dense retrievers on most tasks.

\begin{figure*}[ht]
    \centering
    \includegraphics[width=\textwidth]{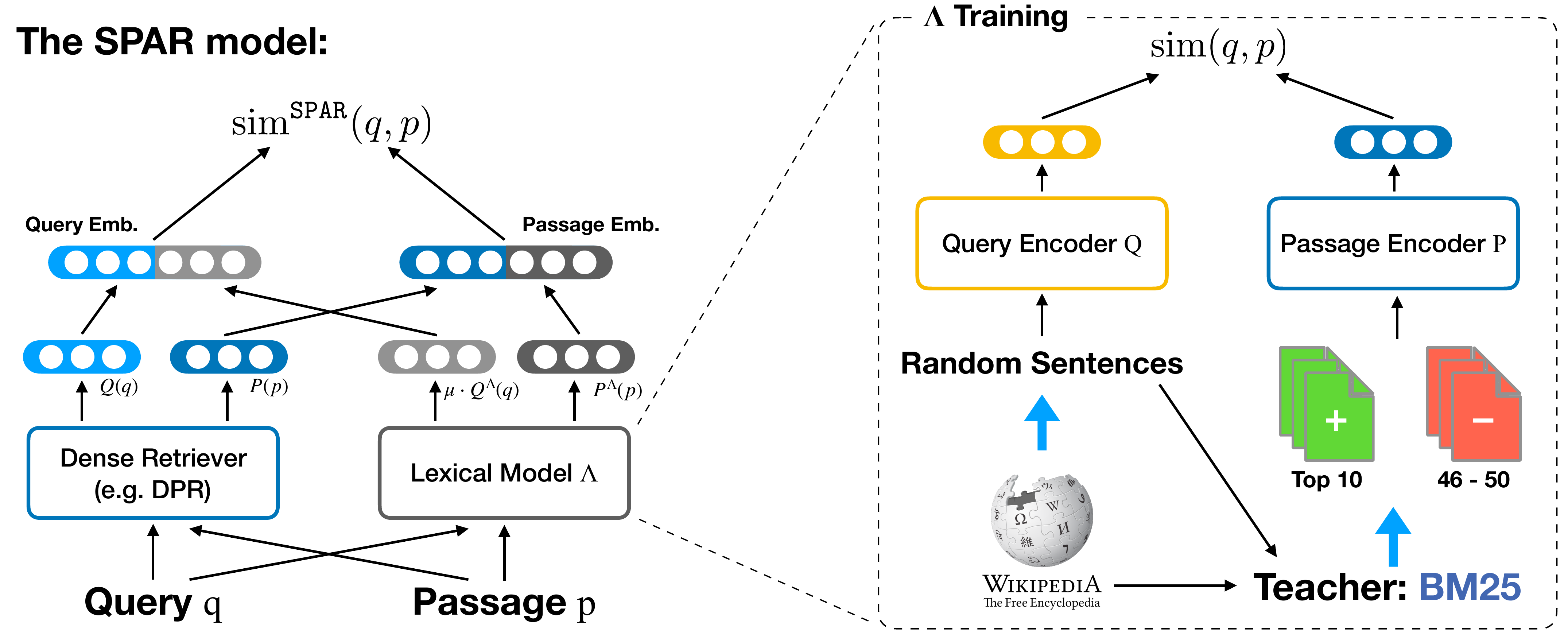}
    \caption{\spar{} augments a dense retriever with a dense \lexmodel{} \lexmodelsymbol{} trained to imitate a sparse teacher retriever. \lexmodelsymbol{} is trained using random sentences as queries with positive and negative passages produced by the teacher. \lexmodelsymbol{} is then combined with a dense retriever via vector concatenation to form a salient-phrase aware retriever.}
    \label{fig:spar_model}
\end{figure*}

With dense and sparse retrievers each having their own distinctive pros and cons, researchers have long aspired to develop retriever models that combine the strengths of both.
This, however, has proven challenging as dense and sparse retrievers are supported by drastically different algorithms and data structures (inverted index~\cite{bialecki2012apache} for sparse and approximate nearest neighbor search~\cite{faiss} for dense).
Most existing research towards this goal extends sparse retrievers with improved representations from neural models~\cite{lin2021brief}.
These methods, nonetheless, still rely on exact matching on a bag of tokens, which arguably cannot fully leverage the representational power of the pre-trained encoders.

The opposite route of \emph{building better dense retrievers with the strengths of sparse models} is much less explored.
In fact, there have been theoretical and empirical studies suggesting that such drawbacks of dense retrievers may be a result of inherent limitations~\cite{luan-etal-2021-sparse,reimers-gurevych-2021-curse}.
In this work, we embark on this underexplored research direction by proposing \spar{} (Fig.~\ref{fig:spar_model}), a \emph{dense} retriever with the lexical matching capacity and out-of-domain generalization of a sparse model.
In particular, we address an important and yet largely unanswered research question: \textbf{Can we train a dense retriever to imitate a sparse one?}
Contrary to previous findings, we show that it is indeed possible to mimic a given sparse retriever (e.g.,~BM25 or UniCOIL~\cite{lin2021brief}) with a dense \lexmodel{} \lexmodelsymbol{}, and we build the \spar{} model by combining \lexmodelsymbol{} with a standard dense retriever (e.g.,~DPR or ANCE).
Despite the long-standing dichotomy between sparse and dense retrievers, we arrive at a simple yet elegant solution of \spar{} by conducting an extensive study to answer two key questions: i) How to train \lexmodelsymbol{} to imitate a sparse retriever (\secref{sec:spar_model:data}) and ii) How to best utilize \lexmodelsymbol{} to build a salient-phrase aware dense retriever (\secref{sec:spar_model:use}).

We evaluate \spar{} on five ODQA datasets (\secref{sec:exp:odqa}) as well as on the MS MARCO~\cite{bajaj2018ms} passage retrieval benchmark (\secref{sec:exp:msmarco}), and show that it outperforms existing dense and sparse retrievers.
We also examine the out-of-domain generalization of \spar{} showing strong \textbf{zero-shot} performance across datasets (\secref{sec:discussion:generalization}), including on the BEIR bechmark~\cite{thakur2021beir} and a recently released dataset of entity-centric questions~\cite{sciavolino2021simple}.
In addition, we conduct analyses of \lexmodelsymbol{} showcasing its lexical matching capability (\secref{sec:discussion:lexical}).

\section{Related Work}\label{sec:relatedwork}

\noindent\textbf{Sparse retrievers} date back for decades and successful implementations such as BM25~\cite{bm25} remain popular to date for its lexical matching capacity and great generalization.
Despite the rapid rise of dense retrievers in recent years, development in sparse retrievers remain active, partly due to the limitations of dense retrievers discussed in \secref{sec:intro}.
Various methods have been proposed to improve term weight learning~\cite{deepct,deepimpact}, address vocabulary mismatch~\cite{doctttttquery} and semantic mismatch~\cite{gao-etal-2021-coil}, \emph{inter alia}.
While most of these methods have been incompatible with dense retrievers, our \spar{} method provides a route for incorporating any such improvement into a dense retriever.

\noindent\textbf{Dense retrievers} employ pre-trained neural encoders to learn vector representations and perform retrieval by using nearest-neighbor search in this dense embedding space~\cite{lee-etal-2019-latent,karpukhin-etal-2020-dense}.
Subsequent works have developed various improvements, including more sophisticated training strategies and using better hard negatives~\cite{xiong2021approximate,qu-etal-2021-rocketqa,maillard-etal-2021-multi,dpr-scale}.
Such improvements are also complementary to the \spar{} approach, which can potentially leverage these more powerful dense retrievers as shown in \secref{sec:exp:msmarco}.

A few recent studies focus on the limitations of current dense retrievers.
\citet{lewis-etal-2021-question} and \citet{liu2021challenges} study the generalization issue of dense retrievers in various aspects, such as the overlap between training and test data, compositional generalization and the performance on matching novel entities.
\citet{thakur2021beir} introduce a new BEIR benchmark to evaluate the zero-shot generalization of retrieval models showing that BM25 outperforms dense retrievers on most tasks.
A different line of research explores using multiple dense vectors as representations which achieves higher accuracy but is much slower~\cite{colbert}.
\citet{lin-etal-2021-batch} further propose a knowledge distillation method to train a standard dense retriever with similar performance of the multi-vector ColBERT model.
More recently, \citet{sciavolino2021simple} create EntityQuestions, a synthetic dataset of entity-centric questions to highlight the failure of dense retrievers in matching key entities in the query.
We evaluate the generalization of \spar{} on BEIR and EntityQuestions in \secref{sec:discussion:generalization}.

\noindent\textbf{Hybrid retrievers} directly combine sparse and dense retrievers, and have been the most commonly used approach to overcome the limitations of a dense or sparse retriever~\cite{DBLP:conf/ecir/GaoDCFDC21,ma2021replication} before this work.
A hybrid system retrieves two separate sets of candidates using a dense and a sparse retriever and rerank them using the hybrid retrieval score.
Compared to a hybrid retriever, \spar{} offers the same performance with a much simpler architecture.
We compare \spar{} with the hybrid models in more details in~\secref{sec:spar_model:spar_vs_hybrid}.

\section{Preliminaries: Dense Retrieval}
\label{sec:background}

In this work, we adopt DPR~\cite{karpukhin-etal-2020-dense} as our dense retriever architecture for learning the \lexmodel{} \lexmodelsymbol{}.
We give a brief overview of DPR in this section and refer the readers to the original paper for more details.

DPR is a bi-encoder model with a \emph{query encoder} and a \emph{passage encoder}, each a BERT transformer~\cite{devlin-etal-2019-bert}, which encodes the queries and passages into $d$-dimensional vectors, respectively.
Passage vectors are generated offline and stored in an index built for vector similarity search using libraries such as FAISS~\cite{faiss}.
The query embedding is computed at run time, which is used to look up the index for $k$ passages whose vectors are the closest to the query representation using dot-product similarity.

DPR is trained using a contrastive objective: given a query and a positive (relevant) passage, the model is trained to increase the similarity between the query and the positive passage while decreasing the similarity between the query and negative ones.
It is hence important to have \emph{hard negatives} (irrelevant passages that are likely confused with positive ones) for more effective training\footnote{DPR uses BM25 to generate hard negatives.}.

We employ the DPR implementation from~\citet{dpr-scale}, which supports efficient multi-node training as well as memory-mapped data loader, both important for the large-scale training of \lexmodelsymbol{}.
For model training, we also adopt their validation metrics of mean reciprocal rank (MRR) on a surrogate corpus using one positive and one hard negative from each query in the development set.
Assuming a set of $N$ dev queries, this creates a mini-index of $2N$ passages, where the MRR correlates well with full evaluation while being much faster.

\section{The \spar{} Model}\label{sec:spar_model}

In this section, we present \spar{}, our salient phrase aware dense retriever.
As illustrated in Figure~\ref{fig:spar_model}, the basic idea of \spar is to first train a dense \lexmodel{} \lexmodelsymbol such that it produces similar predictions to a sparse teacher retriever.
\lexmodelsymbol{} is then combined with a regular dense retriever via vector concatenation.
Although the high-level idea of \spar{} is fairly straightforward, details of the model training process dictate its success in practice, and thus require a careful experimental study. 
To find the best configuration of \spar{}, we conduct pilot experiments using the validation set of NaturalQuestions~\citep[NQ,][]{kwiatkowski-etal-2019-natural} following the ODQA setting~\cite{lee-etal-2019-latent}.
BM25 is adopted as the teacher model for training \lexmodelsymbol{}.
Below we describe the key results on how to successfully train \lexmodelsymbol{} (\secref{sec:spar_model:data}) and on how to best leverage \lexmodelsymbol{} in \spar (\secref{sec:spar_model:use}).

\subsection{Training the \lexmodel{} \lexmodelsymbol{}}\label{sec:spar_model:data}

There are many options to train \lexmodelsymbol{} to imitate the predictions of a sparse retriever, such as mimicking the scores of the sparse retriever with the MSE loss or KL divergence, or learning the passage ranking of the teacher while discarding the scores.
After unsuccessful initial attempts with these methods, we instead choose a very simple approach inspired by the DPR training. 
Recall that to train a DPR model, a \emph{passage corpus} and a set of \emph{training queries} are needed, where each query is associated with one or more \emph{positive} and \emph{hard negative} passages.
To create such data for training \lexmodelsymbol{}, we use the sparse teacher retriever to produce the positive and negative passages.
In particular, for any given query, we run the teacher model to retrieve the top $K$ passages and use the top $n_p$ passages as positives and the bottom $n_n$ as negatives.
After the training data is generated, \lexmodelsymbol{} can be trained using the same contrastive loss as DPR.
To find the best strategy for training \lexmodelsymbol, we experiment with different training queries and different values of $K$, $n_p$ and $n_n$.

We use a similar process to create the \emph{validation} data, and adopt the MRR metric (\secref{sec:background}) for evaluating whether \lexmodelsymbol{} behave similarly to the teacher model.
In particular, we use the questions from the NQ validation set as the target queries and whether a passage is relevant is again determined by the teacher.
For validation, $n_p$ is set to 1, which means only the passage ranked the highest by the teacher is considered positive.
As a result, a higher MRR on this validation data indicates that the predictions of \lexmodelsymbol{} are more similar to the teacher.

\paragraph{Training queries}

As the teacher model can be run on arbitrary queries to generate positive and negative passages, we are not restricted to any annotated data for training \lexmodelsymbol{}.
We experiment with different approaches for constructing such queries, focusing on two potentially helpful properties: i) how similar the queries are to the downstream evaluation queries and ii) the quantity of queries as large-scale training data gives the neural models more exposure to rare words and phrases.
Specifically, we consider three query sets: NQ questions, Wiki sentences and PAQ questions.
NQ questions consists of 59k questions in the training set of NQ, which have the benefit of being similar to the evaluation queries.
Wiki sentences are a large collection of 37 million sentences sampled randomly from the Wikipedia passage corpus. (See \secref{sec:exp:implementation} for how the queries are sampled.)
Finally, PAQ questions are from the PAQ dataset~\cite{lewis2021paq}, a collection of 65 million synthetically generated \emph{probably asked questions} based on Wikipedia, which have both desirable properties.

Based on our experiments, NQ \lexmodelsymbol{} achieves a MRR of 76.5\%, while Wiki \lexmodelsymbol{} attains 92.4\% and PAQ \lexmodelsymbol{} reaches 94.7\%.
This indicates that the size of the data certainly plays an important role, as both Wiki \lexmodelsymbol{} and PAQ \lexmodelsymbol{} outperform NQ \lexmodelsymbol{} by a wide margin.
PAQ \lexmodelsymbol{} achieves the highest MRR among the three options, for it combines the benefits of NQ and Wiki \lexmodelsymbol{}, with large-scale queries that are also similar to the downstream evaluation data.
However, such large-scale synthetic queries are very expensive to obtain and not available for all domains, so we focus our attention on the much cheaper Wiki \lexmodelsymbol{} option, given that the gap between their performance is small.

\paragraph{Number of positive and negative passages}

\begin{figure}[t]
    \centering
    \includegraphics[width=\linewidth]{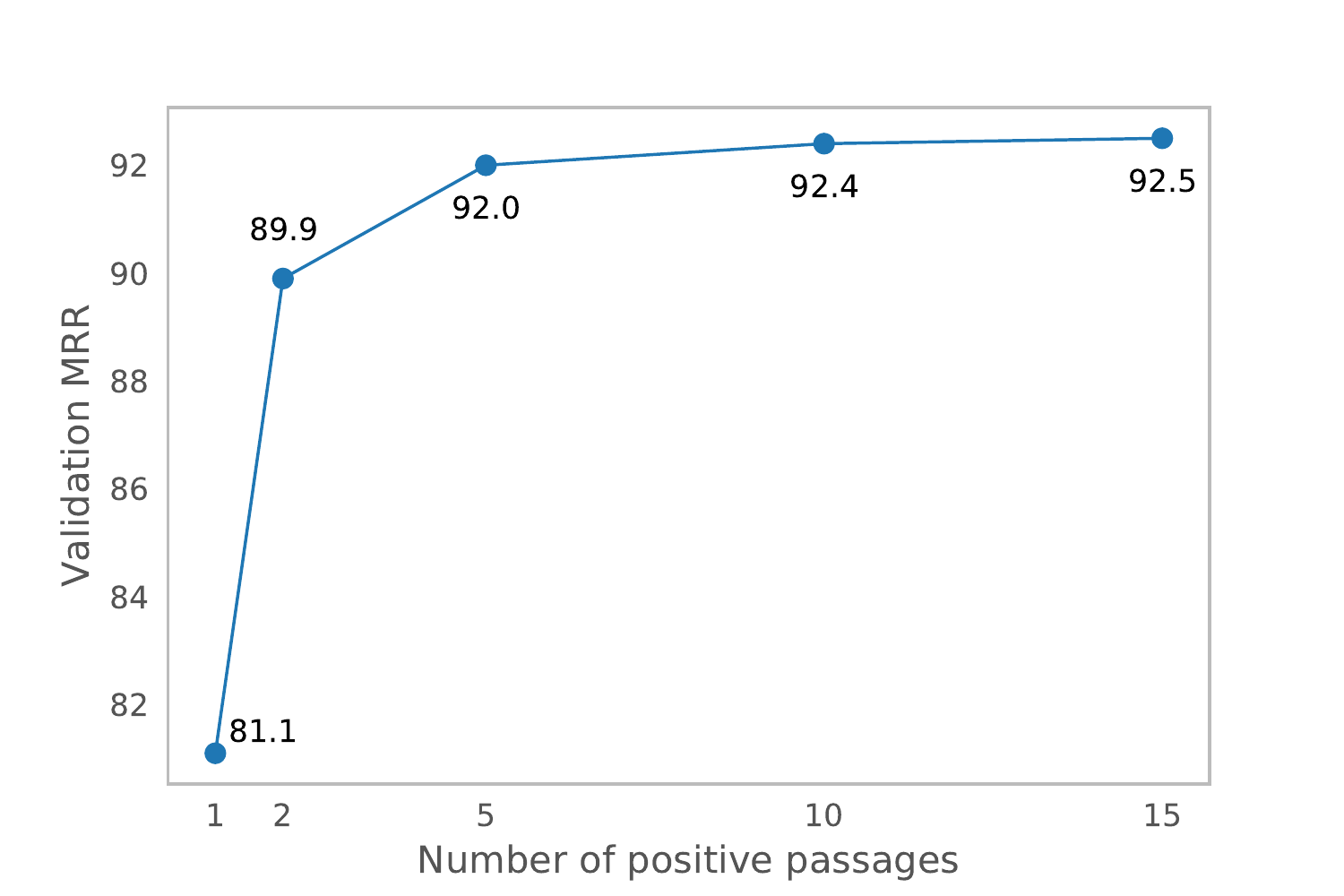}
    \caption{Validation MRR of the Wiki \lexmodelsymbol{} using various numbers of positive passages.}
    \label{fig:num_positive}
\end{figure}

We also experimented with the numbers of positive and negative passages per training query. 
The model performance was not sensitive to the total number of passages retrieved from the teacher model ($K$) as well as the number of negative passages ($n_n$).
However, the number of positives ($n_p$) is more important.
As shown in Figure~\ref{fig:num_positive}, using more than one positive passage is critical to successful training, as $n_p=2$ significantly improves the validation metrics over $n_p=1$.
Further increase in $n_p$ remains helpful, but with a diminished return. As a result, we use $n_p=10$ in our final model.

\subsection{Building \spar{} with \lexmodelsymbol{}}\label{sec:spar_model:use}

With a successfully trained dense lexical model \lexmodelsymbol{}, we next experiment with various approaches to build a salient-phrase aware retriever with it.
All models are trained on NQ training set and we report the Acc@20 and Acc@100 metrics on the test set, which evaluates whether any of the top 20/100 retrieved passages contain the answer to the input question.
We consider two baseline models, DPR (NQ-single) and the hybrid DPR+BM25 model\footnote{We implement DPR+BM25 with Pyserini~\cite{pyserini}; see \secref{sec:spar_model:spar_vs_hybrid} for more discussions on hybrid models.}.
Because \spar{} is built by augmenting DPR with \lexmodelsymbol, it should perform better than the DPR alone.
Moreover, if our dense lexical model is effective, then the final model should achieve a comparable performance as DPR+BM25.

The first method we test is to initialize DPR with the model weights of \lexmodelsymbol (Initialization) for DPR training, with the hope that DPR can inherit the lexical matching capacity from \lexmodelsymbol{}. However, as shown in Table~\ref{tab:lexical_model_use}, the results are only slightly better than the original DPR on Acc@100.
We next test directly combining the vectors of DPR and \lexmodelsymbol, using either summation (Weighted Sum) or concatenation (Weighted Concat), where the weights are tuned using the validation data.
In our experiments, we find that both methods perform well, achieving higher Acc@20 and Acc@100 scores than DPR+BM25. 
Although concatenation performs better, summation has the benefit of not increasing the dimension of the final vectors.
Since both DPR and \lexmodelsymbol{} are dense retrievers, this \emph{post-hoc} combination can be easily done while still using a single FAISS index.

The good performance of Weighted Concat is encouraging, but also triggers two questions. 
First, does the good performance come from the longer embedding size? 
To answer this question, we include the ensemble (weighted concatenation of embeddings) of two independently trained DPR models (2 DPRs) as an additional baseline.
Although it has the same dimension and number of parameters as Weighted Concat, its performance is substantially lower.
Second, can we train a better concatenation model instead of simply combining the vectors of two separately trained models at test time?
We experiment with a joint training approach, in which we concatenate the DPR embeddings with that of a trained \lexmodelsymbol{} during DPR training.
The similarity and loss are computed with the concatenated vector, but we freeze \lexmodelsymbol{}, and only train the DPR encoders as well as the scalar weight for vector concatenation.
The idea is to make DPR ``aware'' of the lexical matching scores given by \lexmodelsymbol{} during its training in order to learn a \spar{} model.
This can also be viewed as training DPR to correct the errors made by \lexmodelsymbol{}.
Somewhat surprisingly, however, this strategy does not work well compared to post-hoc concatenation as shown in Table~\ref{tab:lexical_model_use}.
We hence adopt the Weighted Concat method in our final model.

\begin{table}[t]
    \centering
    \small
    \begin{tabular}{l c cc}
    \toprule
    Acc@$k$ on NQ && @20 & @100 \\
    \midrule
    DPR     && 78.3 & 85.6 \\
    2 DPRs (Weighted Concat)             && 79.8 & 86.4 \\
    DPR + BM25 Hybrid                  && 80.9 & 87.9 \\
    \midrule
    \spar{} (Initialization)     && 78.1 & 86.3 \\
    \spar{} (Joint Training)     && 79.2 & 86.6 \\
    \spar{} (Weighted Sum)        && 81.3 & 88.0 \\
    \spar{} (Weighted Concat)        && \bf 82.2 & \bf 88.3 \\
    \bottomrule
    \end{tabular}
    \caption{Comparison of various methods of leveraging the Wiki \lexmodelsymbol{} in \spar{}: used as initialization for DPR training; combining two trained models with weighted vector sum or concatenation; vector concatenation during DPR training (joint training).}
    \label{tab:lexical_model_use}
\end{table}

\paragraph{Concatenation Weight Tuning}
When concatenating the vectors from two dense retrievers, they may be on different scales, especially across varied datasets.
It is hence helpful to add a weight $\mu$ to balance the two models during concatenation. 
We \emph{add the weight to the query embeddings} so that the offline passage index is not affected by a change of weight.
Specifically, for a query $q$ and a passage $p$, a dense retriever with query encoder $Q$ and passage encoder $P$, as well as a \lexmodelsymbol{} model with $Q^\Lambda$ and $P^\Lambda$, the final query vector in \spar{} is $\left[Q(q),  \mu Q^\Lambda(q)\right]$ while the passage vector being $\left[P(p), P^\Lambda(p)\right]$.
The final similarity score is equivalent to a linear combination of the two model scores:
\begin{align}
\score^{\spar{}}(q,p) &= \left[Q(q),  \mu Q^\Lambda(q)\right]^\intercal \left[P(p), P^\Lambda(p)\right] \nonumber \\
&= \score(q,p) + \mu \cdot \score^\Lambda(q,p)
\label{eqn:sim_score}
\end{align}
Note that unlike hybrid retrievers, the similarity function of \spar{} is an \textbf{exact} hybrid of \lexmodelsymbol{} and the base retriever, achievable in a single FAISS index search, thanks to the fact that both are dense retrievers.
In addition, our decision of adding $\mu$ to the query vectors can potentially support dynamic or query-specific weights without the need to change the index, which we leave for future work.

\noindent\textbf{Our final \spar{} model} is a general framework for augmenting any dense retriever with the lexical matching capability from any given sparse retriever.
We first train \lexmodelsymbol{} using queries from random sentences in the passage collection and labels generated by the teacher model with 10 positive and 5 hard negative passages.
We then combine \lexmodelsymbol{} and the base dense retriever with weighted vector concatenation using weights tuned on the development set.
The passage embeddings can still be generated offline and stored in a single FAISS index and retrieval can be done in the same way as a standard dense retriever.
Further implementation details can be found in \secref{sec:exp:implementation}.

\begin{table*}[ht]
\setlength{\tabcolsep}{0.35em}
\small
\centering
\begin{threeparttable}
\begin{tabular} {l c cc c cc c cc c cc c cc c cc}
\toprule
   &&    \multicolumn{2}{c}{NQ} &&  \multicolumn{2}{c}{SQuAD} && \multicolumn{2}{c}{TriviaQA} &&  \multicolumn{2}{c}{WebQ} &&  \multicolumn{2}{c}{TREC}  && \multicolumn{2}{c}{\bf Average}\\
\cmidrule{3-4}\cmidrule{6-7}\cmidrule{9-10}\cmidrule{12-13}\cmidrule{15-16}\cmidrule{18-19}
Model    && @20  & @100 &&  @20  & @100   && @20  & @100   && @20  & @100   && @20  & @100   && @20  & @100  \\ 
\midrule
({\bf s}) BM25 && 62.9 & 78.3 && 71.1 & 81.8 && 76.4 & 83.2 && 62.4 & 75.5 && 80.7 & 89.9 &&  70.7 & 81.7 \\
\midrule
({\bf d}) Wiki \lexmodelsymbol{} && 62.0 & 77.4 && 67.6 & 79.4 && 75.7 & 83.3 && 60.4 & 75.0 && 79.8 & 90.5 && 69.1 & 81.1 \\
({\bf d}) PAQ \lexmodelsymbol{} && 63.8 & 78.6 && 68.0 & 80.1 && 76.5 & 83.4 && 63.0 & 76.4 && 81.0 & 90.5 && 70.5 & 81.8 \\
\midrule
({\bf d}) DPR-multi && 79.5 & 86.1 && 52.0 & 67.7 && 78.9 & 84.8 && 75.0 & 83.0 && 88.8 & 93.4 && 74.8 & 83.0 \\
({\bf d}) xMoCo\tnote{1} && 82.5 & 86.3 && 55.9 & 70.1 && 80.1 & 85.7 && \bf \underline{78.2} & 84.8 && 89.4 & 94.1 && 77.2 & 84.2 \\
({\bf d}) ANCE\tnote{2} && 82.1 & 87.9 && - & - && 80.3 & 85.2 && - & - && - & - && - & - \\
({\bf d}) RocketQA\tnote{3} && 82.7 & 88.5 && - & - && - & - && - & - && - & - && - & - \\

\midrule
({\bf h}) DPR + BM25\tnote{4} && 82.6 & 88.6 && \bf 75.1 & \bf 84.4 && \bf 82.6 & 86.5 && 77.3 & 84.7 && 90.1 & 95.0 && \bf 81.5 & 87.8 \\

\midrule
({\bf d}) \spar{}-Wiki && \bf \underline{83.0} & \bf \underline{88.8} && \underline{73.0} & 83.6 && \bf \underline{82.6} & 86.7 && 76.0 & 84.4 && 89.9 & 95.2 && \underline{80.9} & 87.7 \\
({\bf d}) \spar{}-PAQ && 82.7 & 88.6 && 72.9 & \underline{83.7} && 82.5 & \bf \underline{86.9} && 76.3 & \bf \underline{85.2} && \bf \underline{90.3} & \bf \underline{95.4} && \underline{80.9} & \bf \underline{88.0} \\
\specialrule{\heavyrulewidth}{\aboverulesep}{\belowrulesep}
\multicolumn{10}{l}{\small\emph{Cross-dataset model generalization (Discussed in \secref{sec:discussion:generalization})}}\\
({\bf d}) \spar{}-MARCO && 82.3 & 88.5 && 71.6 & 82.6 && 82.0 & 86.6 && 77.2 & 84.8 && 89.5 & 94.7 && 80.5 & 87.4 \\

\bottomrule
\end{tabular}

\begin{tablenotes}[para,raggedright]
\item[1] \citep{yang-etal-2021-xmoco}
\item[2] \citep{xiong2021approximate}
\item[3] \citep{qu-etal-2021-rocketqa}
\item[4] \citep{ma2021replication}
\end{tablenotes}

\end{threeparttable}

\caption{Acc@$20$ and 100 for Open-Domain Question Answering. Model types are shown in parentheses (\textbf{d}: dense, \textbf{s}: sparse, \textbf{h}: hybrid). The highest performance is in bold, and the highest among dense retrievers is underlined.}
\label{tab:odqa-results}
\end{table*}

\subsection{Comparing \spar{} with Hybrid Retrievers}\label{sec:spar_model:spar_vs_hybrid}

While most research focuses on improving either dense or sparse retrievers to overcome their drawbacks, people in practice also build hybrid retrievers that directly incorporate both.
In addition to addressing an open research question probing the limits of a dense retriever, \spar{} also has several \emph{practical} advantages over such hybrid models, which is discussed in this section.

\paragraph{Architectural Complexity}
A major advantage of \spar{} is its simplicity as a dense retriever.
Hybrid retrievers need to build and search from two separate indices with different libraries (e.g.~FAISS for dense and Lucene for sparse), and the retrieved passages from the dense and the sparse index then need to be aggregated to form the final results.
\spar{}, on the other hand, can be deployed and used in the same way as any standard dense retriever such as DPR without added complexity.
Passage embeddings can be pre-computed and stored in a single FAISS index, and only a single lookup in the FAISS index is needed for retrieval.
Furthermore, most dense retrieval optimizations such as Product Quantization, HNSW can also be applied to \spar{}.
Last but not least, it is prohibitive to perform an exact hybrid search in hybrid models and challenging to even devise an efficient approximation~\cite{wu2019efficient}.
\spar{} retrieval, however, is equivalent to an \textbf{exact} hybrid of \lexmodelsymbol{} and the base retriever (Eqn.~\ref{eqn:sim_score}), without the need for approximation.

\paragraph{Retrieval Speed and Index Size}
\spar{} has two variants (\secref{sec:spar_model:use}), where the \emph{weighted concat} variant is optimized for accuracy while the \emph{weighted sum} variant has higher efficiency.
With the weighted sum variant, the index size and retrieval speed stays the same as the base dense retriever, making it superior than a hybrid model.

For the weighted concat variant, \spar{} index search takes 20ms/query using a HNSW index (compared to DPR's 10ms).
BM25, in comparison, has a latency of 55ms using the Anserini toolkit~\cite{10.1145/3404835.3462891}, and a hybrid DPR+BM25 retriever has a latency of 58ms assuming DPR and BM25 retrieval can be done in parallel.
For the index size, however, a hybrid model may have an advantage thanks to the small index footprint of BM25.
\spar{}'s index for MS MARCO is 52GB when using weighted concat, which is twice as large as DPR.
A hybrid DPR+BM25 model, instead, has an index size of 27GB as BM25 only takes up 700MB space.

\section{Experiments}\label{sec:exp}

\subsection{Open-Domain Question Answering}\label{sec:exp:odqa}

\begin{table*}[ht]
\setlength{\tabcolsep}{0.35em}
\small
    \centering
    \begin{tabular}{l c ccc}
    \toprule
    && \multicolumn{3}{c}{MS MARCO Dev Set} \\
    \cmidrule{3-5}
     Model          && MRR@10 & R@50 & R@1000 \\
    \midrule
    ({\bf s}) BM25    && 18.7 & 59.2 & 85.7 \\
    ({\bf s}) UniCOIL~\cite{lin2021brief} && 35.2 & 80.7 & 95.8 \\
    \midrule
    ({\bf d}) MARCO BM25 \lexmodelsymbol{}    && 17.3 & 56.3 & 83.1 \\
    ({\bf d}) MARCO UniCOIL \lexmodelsymbol{}    && 34.1 & 82.1 & 97.0 \\
    \midrule
    ({\bf d}) ANCE~\cite{xiong2021approximate}    && 33.0 & 79.1 & 95.9 \\
    ({\bf d}) TCT-ColBERT~\cite{lin-etal-2021-batch}    && 35.9 & - & 97.0 \\
    ({\bf d}) RocketQA~\cite{qu-etal-2021-rocketqa} && 37.0 & 84.7 & 97.7 \\
    \midrule
    ({\bf h}) ANCE + BM25 && 34.7 & 81.6 & 96.9 \\
    ({\bf h}) RocketQA + BM25 && 38.1 & 85.9 & 98.0 \\
    ({\bf h}) ANCE + UniCOIL &&  37.5 & 84.8 & 97.6 \\
    ({\bf h}) RocketQA + UniCOIL && \bf 38.8 & \bf 86.5 & 97.3 \\
    \midrule
    ({\bf d}) \spar{} (ANCE + \lexmodelsymbol{}=MARCO BM25) && 34.4 & 81.5 & 97.1 \\
    ({\bf d}) \spar{} (RocketQA + \lexmodelsymbol{}=MARCO BM25) && 37.9 & 85.7 & 98.0 \\
    ({\bf d}) \spar{} (ANCE + \lexmodelsymbol{}=MARCO UniCOIL) && 36.9 & 84.6 & 98.1 \\
    ({\bf d}) \spar{} (RocketQA + \lexmodelsymbol{}=MARCO UniCOIL) && \underline{38.6} & \underline{86.3} & \bf \underline{98.5} \\
    \specialrule{\heavyrulewidth}{\aboverulesep}{\belowrulesep}
    \multicolumn{5}{l}{\small\emph{Cross-dataset model generalization (Discussed in \secref{sec:discussion:generalization})}}\\
    ({\bf d}) Wiki BM25 \lexmodelsymbol{}    && 15.8 & 50.8 & 78.8 \\
    ({\bf d}) \spar{} (ANCE + \lexmodelsymbol{}=Wiki BM25) && 34.4 & 81.5 & 97.0 \\
    ({\bf d}) \spar{} (RocketQA + \lexmodelsymbol{}=Wiki BM25) && 37.7 & 85.3 & 98.0 \\
    \bottomrule
    \end{tabular}
    \caption{\spar{} results on MS MARCO passage retrieval. We consider several options for \lexmodelsymbol{}, trained with different objectives (BM25 and UniCOIL) and different corpora (MSMARCO and Wikipedia). For ANCE and RocketQA, we use the released checkpoints and our evaluation scripts. We matched public numbers in most cases, but we were unable to reproduce the R@50 and R@1000 reported by RocketQA.}
    \label{tab:msmarco_results_full}
\end{table*}

\paragraph{Datasets}
We evaluate on five widely used ODQA datasets~\cite{lee-etal-2019-latent}: NaturalQuestions~\citep[NQ,][]{kwiatkowski-etal-2019-natural}, SQuAD v1.1~\citep{rajpurkar-etal-2016-squad}, TriviaQA~\citep{joshi-etal-2017-triviaqa}, WebQuestions~\citep[WebQ,][]{berant-etal-2013-semantic} and CuratedTREC~\citep[TREC,][]{trec}.
We follow the exact setup of DPR~\cite{karpukhin-etal-2020-dense}, including the train, dev and test splits, and the Wikipedia passage collection, as well as the accuracy@$k$ (Acc@$k$) metric for evaluation, which is defined as the fraction of queries that has at least one positive passage retrieved in the top $k$.

Table~\ref{tab:odqa-results} presents the main results on ODQA.
For \spar{} models, we report two variants both trained with BM25 as teacher, using the Wiki and PAQ training queries respectively (\secref{sec:spar_model:data}).
The MARCO \lexmodelsymbol{} is to test the model generalization of \spar{}, and will be discussed in \secref{sec:discussion:generalization}.
\spar{} outperforms all state-of-the-art retrievers in the literature, usually by wide margins, demonstrating the effectiveness of our approach.

Another appealing result comes from SQuAD, a dataset on which all previous dense retrievers fail to even get close to a simple BM25 model.
As the SQuAD annotators are given the Wikipedia passage when they write the questions, the lexical overlap between the questions and the passages is hence higher than other datasets.
The poor performance of dense retrievers on SQuAD confirms that dense retrieval struggles at lexical matching.
On the other hand, \spar{} dramatically improves over previous models, achieving an improvement of 13.6 points in Acc@100 over the best existing dense retriever.

PAQ \lexmodelsymbol{} matches the accuracy of the teacher BM25 model, while Wiki \lexmodelsymbol{} performs slightly worse.
The performance gap, however, is smaller in the final \spar{} model.
Both approaches are able to match the performance of the hybrid model, and \spar{}-PAQ is only 0.3\% better on average than \spar{}-Wiki.
This enables us to go with the much cheaper Wiki option for training \lexmodelsymbol{} without sacrificing much of the end performance.

\subsection{MS Marco Passage Retrieval}\label{sec:exp:msmarco}

In this section, we report our experiments on the MS MARCO passage retrieval dataset~\cite{bajaj2018ms}, a popular IR benchmark with queries from the Bing search engine and passages from the web.
Following standard practice, we evaluate MRR@10, Recall@50 and Recall@1000.

To highlight the versatility of our approach, we adopt two base dense retrievers in \spar{}, ANCE and RocketQA.
We further consider two sparse retrievers for training \lexmodelsymbol{}, BM25 and UniCOIL~\cite{lin2021brief}, a recent SoTA sparse retriever, to study whether \spar{} training can imitate a more advanced teacher model.
Similar to the Wiki training queries, we create a MARCO corpus for training \lexmodelsymbol{}.
As the MS MARCO passage collection has fewer passages than Wikipedia, we use all sentences instead of sampling, resulting in a total of 28 million queries.
We also report the performance of the Wiki \lexmodelsymbol{} for model generalization (see \secref{sec:discussion:generalization}).

\begin{table*}[ht]
    \centering
    \small
    \setlength{\tabcolsep}{0.33em}
    \begin{tabular}{l ccccccccccccccc}
    \toprule
    & TC & NF & NQ & HQ & FQ & AA & T2 & Qu & CQ & DB & SD & Fe & CF & SF & \bf Avg. \\
    \midrule
    BM25 & 65.6 & 32.5 & 32.9 & 60.3 & 23.6 & 31.5 & \bf 36.7 & 78.9 & 29.9 & 31.3 & 15.8 & 75.3 & 21.3 & 66.5 & 43.0 \\
    docT5query & 71.3 & 32.8 & 39.9 & 58.0 & 29.1 & 34.9 & 34.7& 80.2 & 32.5 & 33.1 & 16.2 & 71.4 & 20.1 & 67.5 & 44.4 \\
    \midrule
    ANCE & 65.4 & 23.7 & 44.6 & 45.6 & 29.5 & 41.5 & 24.0 & 85.2 & 29.6 & 28.1 & 12.2 & 66.9 & 19.8 & 50.7 & 40.5 \\
    ColBERT & 67.7 & 30.5 & 52.4 & 59.3 & 31.7 & 23.3 & 20.2 & 85.4 & 35.0 & 39.2 & 14.5 & 77.1 & 18.4 & 67.1 & 44.4 \\
    Contriever~\cite{izacard2021unsupervised} & 59.6 & 32.8 & 49.8 & 63.8 & 32.9 & 44.6 & 23.0 & 86.5 & 34.5 & 41.3 & 16.5 & 75.8 & 23.7 & 67.7 & 46.6 \\
    GTR-large~\cite{ni2021large} & 55.7 & 32.9 & 54.7 & 57.9 & \bf 42.4 & 52.5 & 21.9 & 89.0 & 38.4 & 39.1 & 15.8 & 71.2 & \bf 26.2 & 63.9 & 47.2 \\
    GTR-large (our reproduction) & 56.3 & 31.4 & 55.1 & 57.8 & 41.1 & 50.9 & 22.0 & 88.5 & 36.2 & 39.5 & 15.5 & 56.6 & 19.8 & 54.0 & 44.6\\
    \midrule
    \spar{} (ANCE + BM25 \lexmodelsymbol{}) & 68.4 & 27.7 & 47.3 & 53.6 & 32.1 & 45.0 & 28.1 & 86.7 & 33.2 & 32.1 & 14.1 & 72.6 & 23.2 & 59.6  & 44.5 \\
    \spar{} (ANCE + UniCOIL \lexmodelsymbol{}) & \bf 76.4 & 31.8 & 51.1 & 63.0 & 32.3 & 47.2 & 30.3 & 86.9 & 35.8 & 36.1 & 15.5 & 80.2 & 23.5 & 62.6 & 48.1 \\
    \spar{} (GTR + BM25 \lexmodelsymbol{}) & 60.7 & 32.3 & 55.7 & 60.9 & 40.9 & 51.4 & 23.5 & 89.3 & 37.4 & 41.1 & 16.3 & 58.0 & 20.5 & 57.3 & 46.1 \\
    \spar{} (GTR + UniCOIL \lexmodelsymbol{}) & 73.7 & \bf 34.0 & \bf 57.0 & 66.2 & 39.8 & \bf 52.6 & 30.3 & \bf 89.6 & \bf 39.1 & 41.9 & \bf 17.2 & 66.5 & 21.4 & 62.8 & 49.4 \\
    \spar{} (Contriever + BM25 \lexmodelsymbol{}) & 63.0 & 33.7 & 51.3 & 66.4 & 34.1 & 45.9 & 24.9 & 87.5 & 35.8 & \bf 42.8 & 16.9 & 76.9 & 24.8 & \bf 69.5 & 48.1 \\
    \spar{} (Contriever + UniCOIL \lexmodelsymbol{}) & 73.5 & 33.8 & 53.1 & \bf 67.6 & 33.7 & 48.8 & 27.5 & 87.0 & 36.5 & 42.2 & 17.1 & \bf 81.2 & 24.5 & 68.3 & \bf 49.6 \\
    \bottomrule
    \end{tabular}
    \caption{Zero-shot results on BEIR~\cite{thakur2021beir}. All \spar{} models, including the concatenation weights, are trained / tuned on MS MARCO. Dataset Legend: TC=TREC-COVID, NF=NFCorpus, NQ=NaturalQuestions, HQ=HotpotQA, FQ=FiQA, AA=ArguAna, T2=Touch\'e-2020, Qu=Quora, CQ=CQADupStack, DB=DBPedia, SD=SCIDOCS, Fe=FEVER, CF=Climate-FEVER, SF=SciFact.}
    \label{tab:beir_results}
\end{table*}

Table~\ref{tab:msmarco_results_full} illustrates the results, where the sections correspond to sparse retrievers, the \lexmodel{}s, state-of-the-art dense retrievers, various hybrid models, and finally \spar{} and the model generalization experiments.
As demonstrated in Table~\ref{tab:msmarco_results_full}, \spar{} is able to augment ANCE and RocketQA with the lexical matching capacity from either BM25 or UniCOIL, leading to a performance close to the hybrid retriever, and again outperforming all existing dense and sparse retrievers with a MRR@10 of $38.6$.
The fact that \spar{} works with not only DPR and BM25, but other SoTA dense and sparse retrievers makes \spar{} a general solution for combining the knowledge of dense and sparse retrievers in a single dense model.

One interesting phenomenon in both experiments is that while \lexmodelsymbol{} by itself achieves a slightly lower performance than the teacher sparse retriever, the final \spar{} model can reach or beat the hybrid model when combined with the same dense retriever.
One possible reason why \spar{} outperforms the hybrid model may be that \spar{} is able to perform an \emph{exact} ``hybrid'' of two retrievers since both are dense models (Eqn.~\ref{eqn:sim_score}), while the hybrid model relies on approximation.
We leave further investigation in this curious finding to future work.

\subsection{Out-of-Domain Generalization of \spar{}}\label{sec:discussion:generalization}

We now focus on another important topic regarding the generalization of \spar{}.
We have shown that Wiki \lexmodelsymbol{} achieves a similar performance to PAQ \lexmodelsymbol{}, making it often unnecessary to rely on sophisticatedly generated queries for training \lexmodelsymbol{}.
A more exciting finding is that \lexmodelsymbol{} also has great \textbf{zero-shot} generalization to other datasets.

In the last section of Table~\ref{tab:odqa-results} and \ref{tab:msmarco_results_full}, we reported \spar{} performance on ODQA using the \lexmodelsymbol{} model built for MS MARCO and vice versa.
In both directions, \lexmodelsymbol{} has a high zero-shot accuracy, and \spar{}'s performance is close to that using in-domain \lexmodelsymbol{}.
This suggests that \lexmodelsymbol{} shares the advantage of better generalization of a sparse retriever, and it may not be always necessary to retrain \lexmodelsymbol{} on new datasets.

\subsubsection{Zero-shot performance on BEIR}\label{sec:discussion:beir}

We further evaluate zero-shot transfer of \spar{} on the BEIR benchmark~\cite{thakur2021beir}, which consists of a diverse set of 18 retrieval tasks gacross 9 domains\footnote{4 tasks were omitted in our evaluation for license reasons.}.
In particular, following the standard setup, all models are trained on MS MARCO and tested on the BEIR benchmarks.
Therefore, we adopt the MARCO \lexmodelsymbol{} models, and combine them with various dense retrievers trained on MS MARCO to form \spar{} models.
As shown in Table~\ref{tab:beir_results}, \spar{} achieves a new state-of-the-art overall performance, and performs the best on 11 out of 14 datasets.
Regardless of the choice of the base dense retriever, adding either the BM25 or UniCOIL \lexmodelsymbol{} can consistently and substantially boost the performance of the base retriever, even for very recent SoTA models such as Contriever~\cite{izacard2021unsupervised} and GTR~\cite{ni2021large}.

\subsubsection{\spar{} on EntityQuestions}\label{sec:discussion:eq}

\begin{table}[ht]
    \small
    \centering
    \begin{tabular}{l c cc}
    \toprule
    Acc@$k$ on EQ        && Acc@20 & Acc@100 \\
    \midrule
    ({\bf d}) DPR     && 56.6 & 70.1 \\
    ({\bf s}) BM25    && 70.8 & 79.2 \\
    ({\bf h}) DPR + BM25    && 73.3 & \bf 82.3 \\
    \midrule
    ({\bf d}) Wiki \lexmodelsymbol{} && 68.4 & 77.5 \\
    ({\bf d}) PAQ \lexmodelsymbol{} && 69.4 & 78.6 \\
    \midrule
    ({\bf d}) \spar{}     && 73.6 & 81.5 \\
    ({\bf d}) \spar{}-PAQ     && \bf 74.0 & 82.0 \\
    \bottomrule
    \end{tabular}
    \caption{Zero-shot performance on the EntityQuestions~\cite{sciavolino2021simple} dataset. We report micro-average instead of macro-average in the original paper.}
    \label{tab:entityquestions}
\end{table}

A concurrent work~\cite{sciavolino2021simple} also identifies the lexical matching issue of dense retrievers, focusing specifically on entity-centric queries.
They create a synthetic dataset containing simple entity-rich questions, where DPR performs significantly worse than BM25.
In Table~\ref{tab:entityquestions}, we evaluate \spar{} on this dataset in a zero-shot setting without any re-training, other than tuning the concatenation weight on the development set.
The result further confirms the generalization of \spar{}.
\lexmodelsymbol{} transfers much better to this dataset than DPR, achieving a slightly lower performance than BM25.
When \lexmodelsymbol{} is combined with DPR,  \spar{} achieves a higher Acc@20 than the hybrid model, and overall an improvement of 17.4 points over DPR.

\subsection{Implementation Details}\label{sec:exp:implementation}

For the Wiki training queries for \lexmodelsymbol{}, we randomly sample sentences from each passage (following the DPR passage split of Wikipedia) following a pseudo-exponential distribution while guaranteeing at least one sentence is sampled from each passage.
The pseudo-exponential distribution would select more sentences in the first few passages of each Wikipedia document, as they tend to contain more answers, resulting in a collection of 37 million sentences (queries) out of 22M passages.\footnote{We did not extensively experiment with sampling strategies, but one preliminary experiment suggested that sampling uniformly may have worked equally well.}
For the MS MARCO passage collection, we use all sentences as training queries without sampling, leading to a total of 28M queries out of 9M passages.

We train \lexmodelsymbol{} for 3 days on 64 V100 GPUs with a per-GPU batch size of 32 and a learning rate of $\expnumber{3}{-5}$ (roughly 20 epochs for Wiki \lexmodelsymbol{} and MARCO \lexmodelsymbol{}, and 10 epochs for PAQ \lexmodelsymbol{}).
The remaining hyper-parameters are the same as in DPR, including the BERT-base encoder and the learning rate scheduler.
For Wiki and PAQ \lexmodelsymbol{}, we use NQ dev as the validation queries, and MS MARCO dev for MARCO \lexmodelsymbol{}.
For the dense retrievers used in \spar{}, we directly take the publicly released checkpoints without re-training to combine with \lexmodelsymbol{}.
We use Pyserini~\cite{pyserini} for all sparse models used in this work including BM25 and UniCOIL.

For tuning the concatenation weights $\mu$, we do a grid search on $[0.1, 1.0]$ (step size 0.1) as well as their reciprocals, resulting in a total of 19 candidates ranging from 0.1 to 10.
The best $\mu$ is selected using the best R@100 for ODQA (\secref{sec:exp:odqa}) and MRR@10 for MS MARCO (\secref{sec:exp:msmarco}) on the development set for each experiment.

\section{Does \lexmodelsymbol{} Learn Lexical Matching?}\label{sec:discussion:lexical}
In this section, we verify whether \lexmodelsymbol{} actually learns lexical matching with a series of analyses.

\subsection{Rank Biased Overlap with BM25}\label{sec:discussion:overlap}

\begin{table}[ht]
    \small
    \centering
    \begin{tabular}{l ccc}
    \toprule
    RBO w/ BM25   & NQ & SQuAD & Trivia \\
    \midrule
    DPR             & .104 & .078 & .170 \\
    Wiki \lexmodelsymbol{} & .508 & .452 & .505 \\
    PAQ \lexmodelsymbol{} & .603 & .478 & .527 \\
    \bottomrule
    \end{tabular}
    \caption{Rank Biased Overlap~\citep[RBO,][]{rbo} between BM25 and various dense retrievers on the dev set. We use the standard $p=0.9$ in RBO.}
    \label{tab:prediction_overlap}
\end{table}

We first directly compare the predictions of \lexmodelsymbol{} against BM25.
As shown in Table~\ref{tab:prediction_overlap}, the prediction of DPR and BM25 are dramatically different from each other, with a RBO of only $0.1$, which is a correlation measure between partially overlapped ranked lists.
In contrast, \lexmodelsymbol{} achieves a much higher overlap with BM25 of around 0.5 to 0.6.

\subsection{Token-shuffled queries}\label{sec:discussion:shuffled}

\begin{table}[ht]
    \small
    \setlength{\tabcolsep}{0.4em}
    \centering
    \begin{tabular}{l c cc c cc c cc}
    \toprule
    Model   && \multicolumn{2}{c}{Original} && \multicolumn{2}{c}{Shuffled} && \multicolumn{2}{c}{$\Delta$}\\
    \cmidrule{3-4}\cmidrule{6-7}\cmidrule{9-10}
            && @20 & @100 && @20 & @100 && @20 & @100 \\
    \midrule
    DPR     && 77.4 & 84.7 && 69.4 & 80.1 && 8.0 & 4.6 \\
    BM25    && 62.3 & 76.0 && 62.3 & 76.0 && 0.0 & 0.0 \\
    Wiki \lexmodelsymbol{} && 60.9 & 74.9 && 60.8 & 74.9 && 0.1 & 0.0 \\
    PAQ \lexmodelsymbol{} && 62.7 & 76.4 && 62.6 & 76.2 && 0.1 & 0.2 \\
    \bottomrule
    \end{tabular}
    \caption{Lexical matching stress test on NQ Dev, using token-shuffled questions. \lexmodelsymbol{} is order agnostic and maintains its performance on shuffled queries.}
    \label{tab:nq_shuffled}
\end{table}

Next, we inspect the lexical matching capacity of \lexmodelsymbol{} in an extreme case where the order of tokens in each question is randomly shuffled.
Table~\ref{tab:nq_shuffled} indicates that the performance of DPR drops significantly on this token-shuffled dataset, while the bag-of-word BM25 model remains completely unaffected.
On the other hand, both Wiki \lexmodelsymbol{} and PAQ \lexmodelsymbol{} remain highly consistent on this challenge set, showing great robustness in lexical matching.

\subsection{Hybrid \spar{} + BM25 model}

\begin{table}[ht]
    \small
    \centering
    \begin{tabular}{l c cc}
    \toprule
    Acc@$k$ on NQ        && @20 & @100 \\
    \midrule
    ({\bf d}) DPR     && 79.5 & 86.1 \\
    ({\bf h}) DPR + BM25    && 82.6 & 88.6 \\
    \midrule
    ({\bf d}) \spar{}-Wiki     && 83.0 & 88.8 \\
    ({\bf h}) \spar{}-Wiki + BM25 && 82.9 & 88.9 \\
    \midrule
    ({\bf d}) \spar{}-PAQ     && 82.7 & 88.6 \\
    ({\bf h}) \spar{}-PAQ + BM25 && 82.7 & 88.8 \\
    \bottomrule
    \end{tabular}
    \caption{The \spar{}+BM25 model yields minimal gains over \spar{}, indicating that \spar{} does well in lexical matching and performs similarly to a hybrid model.}
    \label{tab:spar_hybrid}
\end{table}

To confirm that \spar{} improves DPR's performance by enhancing its lexical matching capability, we add the real BM25 to \spar{} to create a hybrid model.
As demonstrated in Table~\ref{tab:spar_hybrid}, adding BM25 to \spar{} only results in minimal gains, which indicates that \spar{} renders BM25 almost completely redundant and supports our main claim.

\section{Conclusion}\label{sec:conclusion}

In this paper, we propose \spar{}, a salient-phrase aware dense retriever, which can augment any dense retriever with the lexical matching capacity and out-of-domain generalization from a sparse retriever.
This is achieved by training a dense \lexmodel{} \lexmodelsymbol{} to imitate the behavior of the teacher sparse retriever, the feasibility of which remained unknown until this work.
We show that \spar{} outperforms previous state-of-the-art dense and sparse retrievers, matching or even exceeding more complex hybrid systems, on various in-domain and out-of-domain evaluation datasets.

For future work we plan to explore if a dense retriever can be trained to learn lexical matching directly without relying on a teacher model.
This way, we can avoid imitating the errors of the sparse retriever, and devise new ways of training dense retrievers that can potentially surpass hybrid models.
Moreover, there are several intriguing findings in this work that may warrant further study, such as why \spar{}'s Acc@$k$ improves relatively to the hybrid model as $k$ increases, and why joint training is less effective than post-hoc vector concatenation.

\section*{Limitations}

There is a trade-off between accuracy and efficiency when considering the two variants of \spar{} (Section~\ref{sec:spar_model:use}).
The Weighted Concat variant, the main focus of this paper, gives higher accuracy but results in longer query and passage embeddings.
This in turn increases the index size and the retrieval time.
On the other hand, the Weighted Sum variant does not increase the embedding size, but achieves a lower accuracy compared to the Weighted Concat variant as shown in Table~\ref{tab:lexical_model_use}.

As mentioned in Section~\ref{sec:conclusion}, \spar{} relies on a sparse teacher model to learn the \lexmodel \lexmodelsymbol.
It is an intriguing direction for future work to explore whether we can learn \spar{} from scratch without the help of a teacher retriever.

\bibliography{anthology,custom}
\bibliographystyle{acl_natbib}

\end{document}